\documentclass[sigconf]{acmart}

\usepackage{url}
\usepackage{multirow}

\usepackage{amsmath}

\AtBeginDocument{%
  \providecommand\BibTeX{{%
    \normalfont B\kern-0.5em{\scshape i\kern-0.25em b}\kern-0.8em\TeX}}}
    
\setcopyright{acmlicensed}
\copyrightyear{2022}
\acmYear{2022}
\setcopyright{rightsretained}
\acmConference[GECCO `22 Companion]{Genetic and Evolutionary Computation Conference Companion}{July 9--13, 2022}{Boston, MA, USA}
\acmBooktitle{Genetic and Evolutionary Computation Conference Companion (GECCO '22 Companion), July 9--13, 2022, Boston, MA, USA}
\acmDOI{10.1145/3520304.3529037}
\acmISBN{978-1-4503-9268-6/22/07}
\acmPrice{15.00}

\begin{document}

\title[Evolving Unpredictable Halting in Continuous CA]{Selecting Continuous Life-Like Cellular Automata for Halting Unpredictability: Evolving for Abiogenesis}

\author{Q. Tyrell Davis}
\email{qdavis@uvm.edu}
\affiliation{%
  \institution{University of Vermont}
  \streetaddress{205 Farrell Hall}
  \city{Burlington}
  \state{Vermont}
  \country{USA}
  \postcode{05405}
}
\author{Josh Bongard}
\email{jbongard@uvm.edu}
\affiliation{%
  \institution{University of Vermont}
  \streetaddress{205 Farrell Hall}
  \city{Burlington}
  \state{Vermont}
  \country{USA}
  \postcode{05405}
}
%
\renewcommand{\shortauthors}{Davis and Bongard}

\begin{abstract}

Substantial efforts have been applied to engineer CA with desired emergent properties, such as supporting gliders. Recent work in continuous CA has generated a wide variety of compelling bioreminiscent patterns, and the expansion of CA research into continuously-valued domains, multiple channels, and higher dimensions complicates their study. In this work we devise a strategy for evolving CA and CA patterns in two steps, based on the simple idea that CA are likely to be complex and computationally capable if they support patterns that grow indefinitely as well as patterns that vanish completely, and are difficult to predict the difference in advance. The second part of our strategy evolves patterns by selecting for mobility and conservation of mean cell value. We validate our pattern evolution method by re-discovering gliders in 17 of 17 Lenia CA, and also report 4 new evolved CA and 1 randomly evolved CA that support novel evolved glider patterns. The CA reported here share neighborhood kernels with previously described Lenia CA, but exhibit a wider range of typical dynamics than their Lenia counterparts. Code for evolving continuous CA is made available under an MIT License \footnote{\url{https://github.com/rivesunder/yuca}}.

\end{abstract}

%



\maketitle

\section{Introduction}

The anthropic principle has many variants in two categories: those predicated on the universe being somehow fine-tuned to support life (famously espoused by Barrow and Tipler \citep{barrow1986}), and the incorporation of selection bias into reasoning about the universe, described originally by Carter \citep{carter1983}. We can consider similar perspectives when it comes to examining life-like capabilities in cellular automata (CA). 

Von Neumann's universal constructor CA was carefully designed to support universal computation and construction \citep{vonneumann1966}, analogous to the fine-tuning of strong anthropic principle variants. John H. Conway had the opposite idea in developing his Game of Life. Diligent engineering is not necessary: complexity alone is enough for a good chance of being universal. Among objectives motivating Conway's search for complexity was that future states should be difficult to predict \citep{gardner1970, schleicher2013}. Life follows a simple set of birth and survival rules on a binary rectilinear grid. A cell with exactly 3 live neighbors and state 0 becomes 1, or is born. Cells with state 1, and 2 or 3 active neighbors, stay in state 1, or survive. All other cells transition to or stay in state 0. Life-like CA rules are typically written in the Bx/Sy format, \textit{e.g.} the rules of Life are B3/S23. 

A seminal discovery made early in the development of Conway's Life was a minimal mobile pattern called a glider \citep{berlekamp2004}\footnote{In this work we use the term glider to refer to mobile patterns in general, also called spaceships, as well as the original reflex glider from Life.}. Gliders quickly became the cornerstone of computational universality and building computing machines in Life and other CA, and glider discovery has been an important focus of CA research. 

In this work we evolve continuous CA rules for being difficult to predict halting (\textit{i.e.} all cells go to 0), after a number of update steps. We also evolve via selection for equal capability for simple halting and persistence. A second stage of evolution rewards mobile patterns in a given CA, as gliders are useful for building meaningfully computational interactions and historically have been a proxy for complexity and universality in CA.

We validate our glider evolution algorithm by re-discovering gliders in previously described CA from the Lenia framework \citep{chan2019}), and also report several new CA rule combinations that support moving patterns.

\if{0}
    \begin{figure}
    \begin{center}
      \includegraphics[width=0.35\textwidth]{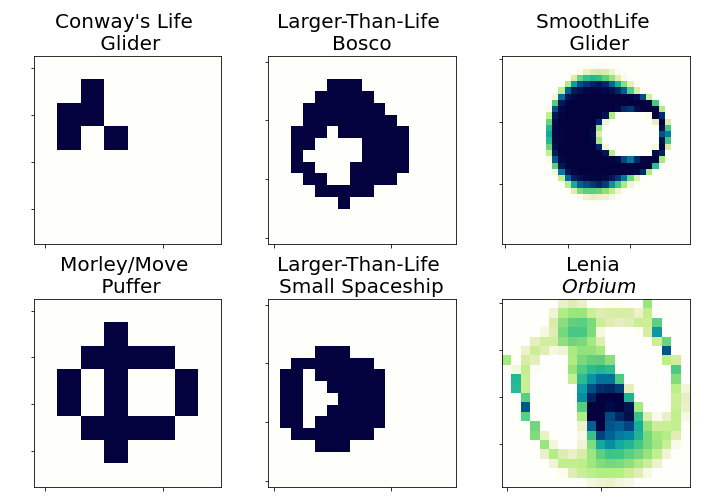}
      \caption{Examples of dynamic, mobile patterns from different cellular automata frameworks. The small glider and common puffer run in Conway's Life and the Life-like Morley ruleset (B368/S245), Bosco and small spaceship are resident in Larger-Than-Life \citep{evans2001}, the top right glider in SmoothLife, and {\it Orbium} is found in a range of variations in the Lenia CA framework.}
      \label{fig:patterns}
    \end{center}
    \end{figure}
\fi

\section{Background}

\subsection{Growth, Decay, and Complexity}

\label{section:bg}

Significant efforts have gone into categorizing or finding specific characteristics of CA systems that underlie complexity and universal computation. These efforts seek to define what makes a complex system `interesting' and to predict whether they are capable of universal computation by developing objective metrics or practical heuristics 

Wolfram proposed a classification scheme for 1D CA with subjective criteria intended to capture complexity and universality \citep{wolfram1983}, also applied to 2D CA \citep{packard1985}. {\bfseries Class I} CA progress to a homogeneous state, typically cells either all become quiescent or all active with a uniform value. {\bfseries Class II} CA settle into static or oscillating pattern equilibria. {\bfseries Class III} CA continuously generate chaotic patterns, usually with relatively uniform statistical characteristics. {\bfseries Class IV} CA generate patterns with complex behavior.

Wolfram's classification system is subjective and many interesting and universal CA fall outside of Class IV \citep{eppstein2010, conwaylifeforum}. 
Eppstein proposed alternative heuristics for predicting universality based on simultaneous support for patterns that are {\itshape mortal}, {\it i.e.} support patterns that vanish, and {\itshape fertile}, {\it i.e.} support patterns that escape a bounding box\citep{eppstein2010}. A simple set of rules that supports patterns that disappear as well as patterns that grow indefinitely was cited as one of the original objectives in developing Life in Gardner's 1970 article, and in later interviews with John H. Conway \citep{gardner1970, schleicher2013}. We use vanishing and growth as the basis for fitness in our evolutionary approach, approximating our real goal: complex CA that support gliders. 

\subsection{Why Gliders?}

\begin{figure} 
\begin{center}
  \includegraphics[width=0.325\textwidth]{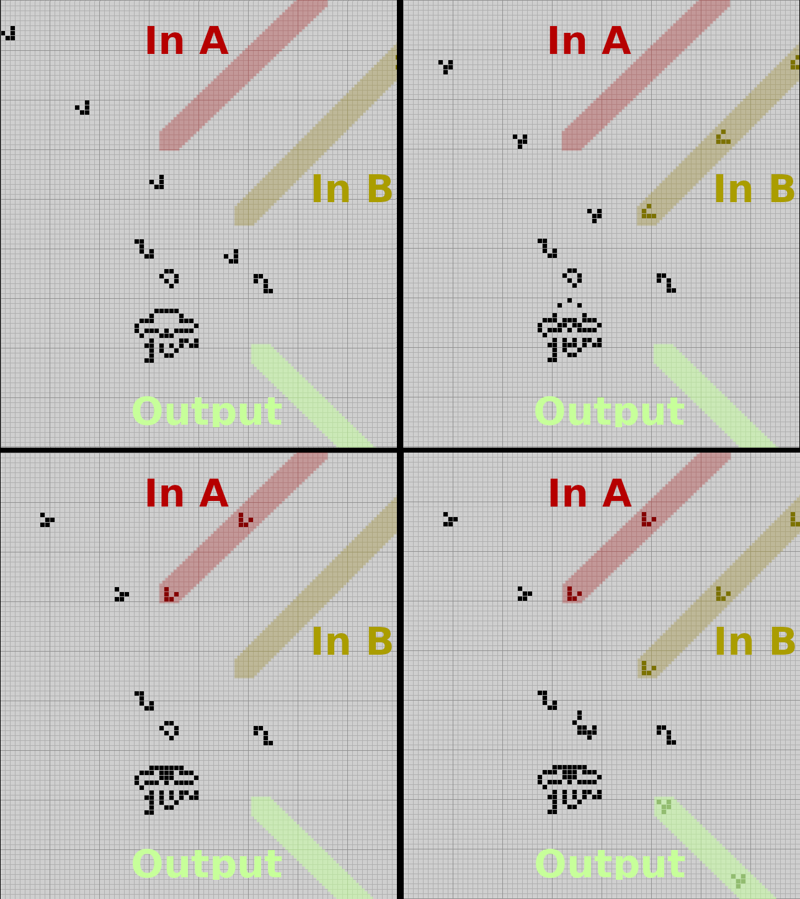}
  \caption{Logical AND gate from Life glider streams. Clockwise from top left: input 00 yields output 0, input 01 yields output 0, input 10 yields output 0, and input 11 yields output 1. This patterns was implemented in Golly \citep{golly}. The pattern in the lower left is simply a reflector that orients the output stream.}
  \label{fig:and_gate}
\end{center}
\end{figure} 

An early discovery that solidified Conway's Life as an interesting automaton was the glider \citep{berlekamp2004}. Gliders can be thought of as elementary units of information: a train of gliders encodes a stream of digital information. Glider collisions can carry out computations and make up the fundamental AND, OR, and NOT logic gates  \citep{gardner1983, berlekamp2004}, such as the AND gate in Figure \ref{fig:and_gate}. As the authors of {\itshape Winning Ways} put it, with basic logic gates and wires defined: 

\begin{quote}
``From here on it's just an engineering problem to construct an arbitrarily large finite (and very slow!) computer. Our engineer has been given the tools---let him finish the job!" \citep{berlekamp2004}
\end{quote}



Computation is an interesting and useful characteristic of complex systems like Life and, as implied in the name, the processes of life have much in common with engineered computation \citep{kempes2017}. Living systems must sense, act on, and copy internal and/or external information in order to be successful. Modern computational resources have facilitated large scale simulation of continuous CA that produce evocative bioreminiscent patterns, at the expense of being somewhat more complicated than their discrete antecedents. 

\subsection{Continuous Cellular Automata}



In addition to Life, CA systems have been developed with larger neighborhoods \citep{evans2001, pivato2007}, higher dimensions \citep{bays1987, chan2020}, and many other extensions. While continuous CA have been developed and applied for modeling tasks for several decades \citep{rucker2003, ruckerweb, rafler2011}, recent work has produced a plethora of bioreminiscent and aesthetically pleasing dynamic patterns, particularly in the Lenia framework \citep{chan2019, chan2020}. 

Parsimony suggests we should prioritize minimally-complicated CA systems that still fulfill desired objectives. Life is by some measures the simplest 2D CA in its class \citep{pena2021}, and is Turing complete \citep{berlekamp2004, rendell2011}, as is the 1D elementary CA 110 \cite{cook2004}. Against the simplicity and complexity of these precedents, do continuous CA offer novel capabilities, or merely appeal to human pareidolia and aesthetic sensibilities? There may indeed be a payoff to the complicated-ness of continuous CA and their successors: self-organizing intelligent agents, fully-embodied in self-consistent simulation. 

Recent work made a nascent demonstration of such self-organizing agents. Authors used gradient descent to train continuous CA update rules, generating robust mobile patterns that survive interactions with immutable obstacles \citep{hamon2022}. Simulated environments are typically distinct from agents in reinforcement learning and evolutionary optimization, even when the environment and the agent are both modeled as different types of CA as in \citep{davis2021}. 
Future work stemming from continuous CA may find systems that exhibit autopoietic selection and robustness (and eventually learning) with no externally imposed evolution or gradient-based optimization. 



\section{Methods}

\subsection{Glaberish Framework}

We use a continuous CA framework called \textit{Glaberish} \citep{davis2022}. Based on Lenia, Glaberish extends Lenia's update function by splitting the single growth function $G$ from Lenia into genesis and persistence functions $G_{gen}$ and $P$, respectively, analogous to the Birth and Survival rules in Life-like CA \citep{gardner1970}. The Glaberish update is shown in Equation \ref{eqn:glaberish}.

\begin{equation}
    A_{t + dt} = \rho \left( A_t + dt \cdot [ (1 - A_t) \cdot G_{gen}(n) + A_t \cdot P(n) ] \right)
    \label{eqn:glaberish}
\end{equation} 

Where $A_t$ is the grid state at time $t$, $n$ is the result of a neighborhood convolution $K*A_t$, $dt$ is the step size, and $\rho$ is a squashing or clipping function that keeps cell values between 0 and 1.

We also consider several CA from the Lenia framework. Lenia differs from Glaberish in that the update is a single growth function, regardless of current cell value. The equation for the Lenia update is shown in Equation \ref{eqn:lenia}.

\begin{equation}
    A_{t + dt} = \rho \left( A_t + dt \cdot G(n) \right)
    \label{eqn:lenia}
\end{equation} 

\subsection{Evolving Halting Unpredictability in Continuously-Valued Cellular Automata}

\begin{figure} 
\begin{center}
  \includegraphics[width=0.45\textwidth]{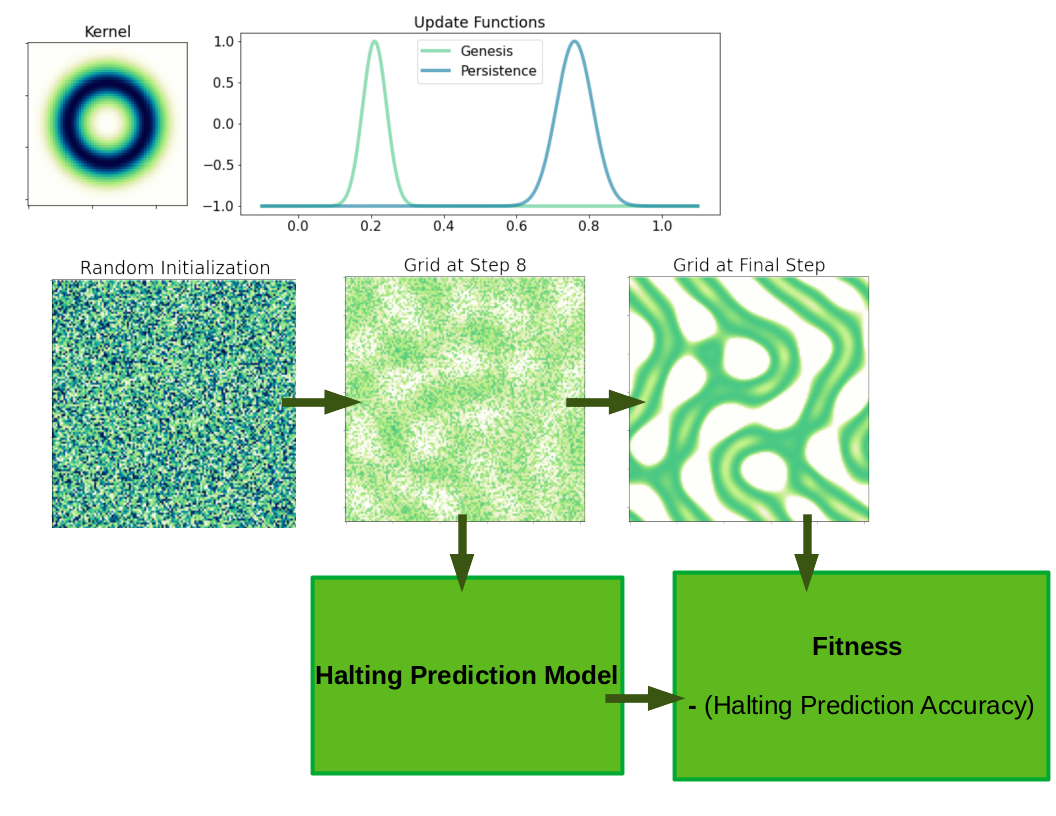}
  \caption{Selecting for halting unpredictability to search for complex systems capable of supporting life-like patterns.}
  \label{fig:teaser}
\end{center}
\end{figure} 

Our approach evolves CA update rules (while keeping a static neighborhood) based on the inability of a trio of convolutional networks to accurately predict whether a CA grid state will settle to a quiescent state (all cells have value 0) or remain active after a number of update steps. To achieve this, we wrapped training/validation of convolutional neural networks in a covariance matrix adaptation evolutionary algorithm \citep{hansen2016}. The (negative) average accuracy of 3 trained models is the fitness (Equation \ref{eqn:halting}). 

\begin{equation}
    max_\theta\left[\mathbb{E} \left( - \frac{1}{N} \sum_n^N(\mathcal{J}(f_{\phi_n}(x), \hat y)) \right)\right]
    \label{eqn:halting}
\end{equation}

Where $\mathcal{J}$ is a function that returns halting prediction accuracy for halting predictions $f_{\phi_n}(x)$ with respect to the final grid states $\hat y$. A cartoon representation of halting unpredictability evolution is shown in Figure \ref{fig:teaser}.

We also implemented a simple version of CA halting evolution. Simple halting evolution has no inner prediction training loop and fitness is based simply on mean-squared error between the proportion of end-point CA grids with nonzero cell values and a target proportion of 0.5. 

\subsection{Evolving Glider Patterns Under Cellular Automata Rule Sets}

\begin{figure}
\begin{center}
  \includegraphics[width=0.45\textwidth]{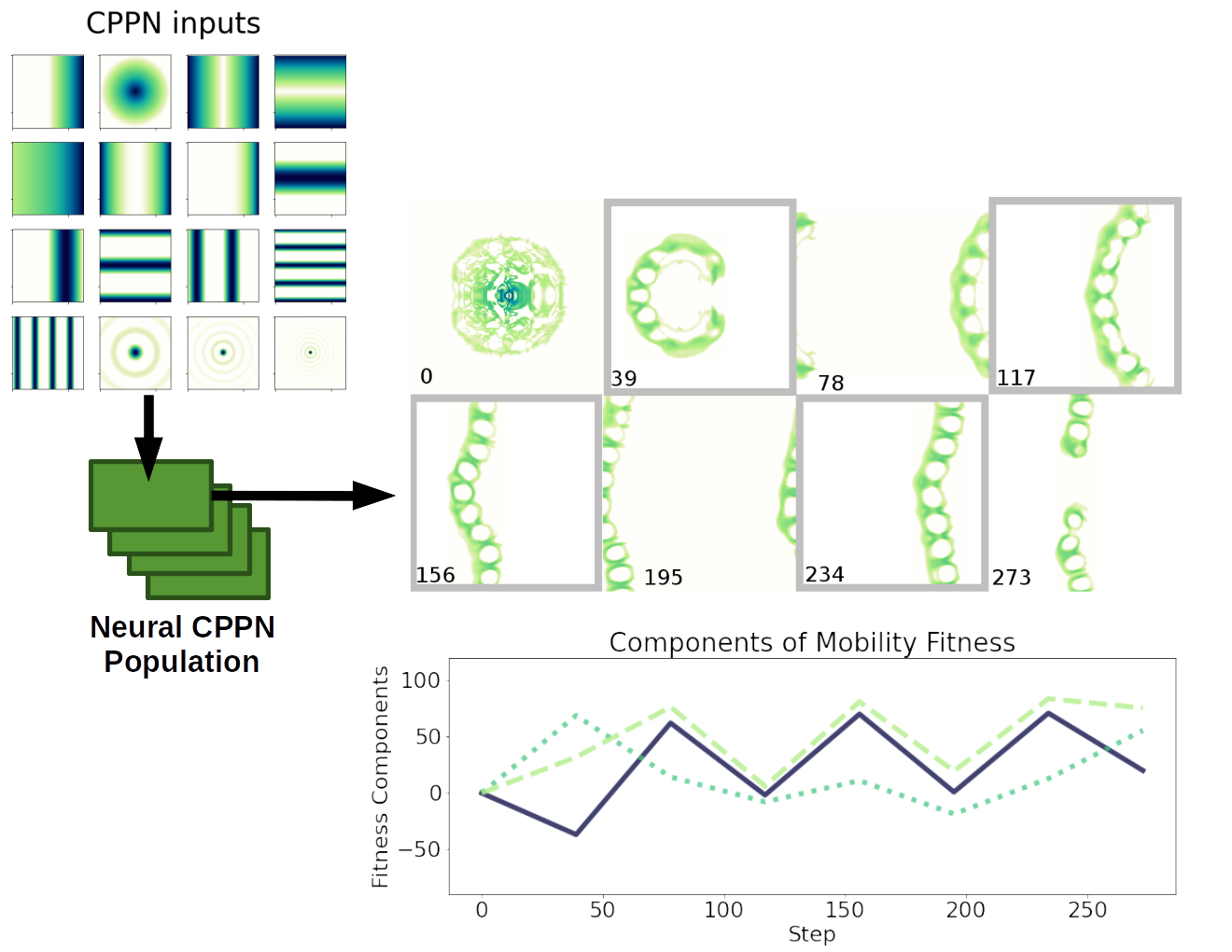}
  \caption{Pattern evolution with mobility-based fitness. Solid line is fitness, dashed line is the motility componet, and dotted line is the homeostasis component.}
  \label{fig:cppn_flow}
\end{center}
\end{figure}

We evolved a population of compositional pattern-producing networks (CPPNs) \citep{stanley2007} as synthesis patterns. Fitness is designed to reward motility, homeostasis, and survival, and is composed of a positive reward for displacement in center of mass (motility), penalized for changes in average cell values (homeostasis), and severely penalized for patterns that disappear entirely (survival). Figure \ref{fig:cppn_flow} shows a graphic representation of pattern evolution. 

\section{Results \& Discussion}

\begin{table}[h]
\center{
\begin{tabular}{c c c}\hline
Name & Origin & $(\mu, \sigma)$ \\ \hline\hline



{\itshape Orbium} & Lenia \citep{chan2019}  & (0.150, 0.0150)  \\


{\itshape P. s. labens} & Lenia \citep{chan2019} & (0.330, 0.0462)  \\


{\itshape S. valvatus} & Lenia \citep{chan2019}  & (0.292, 0.0486)  \\

{\itshape D. valvatus} & Lenia \citep{chan2019}  & (0.337, 0.0595)  \\



{\itshape H. natans}* & Lenia \citep{chan2019} & (0.260, 0.0360)  \\





\hline

\textbf{s7}* & Simple evo. & $(0.0420, 0.00490)_g$ \\
 & &  $(0.261, 0.0292)_p$   \\ \hline
\textbf{s613}* & Pred. evo. & $(0.0621, 0.00879)_g$  \\
 & & $(0.215, 0.0369)_p$   \\ \hline
\textbf{s11}* & Simple evo. & $(0.0761, 0.0107)_g$ \\
& & $(0.260, 0.0303)_p$   \\ \hline
\textbf{s643}* & Simple evo. & $(0.0670, 0.0101)_g$  \\
 & &  $(0.248, 0.0186)_p$   \\ \hline
\textbf{s113}* & Random evo. & $(0.266, 0.0382)_g$  \\
 & & $(0.289, 0.0215)_p$   \\
\end{tabular}
}
\caption{Example CA supporting CPPN-mediated evolved gliders. Update functions are Gaussians with peaks at $\mu$ and width $\sigma$. Lenia CA have a single update function, while CA evolved in this project are based on the Glaberish framework with update functions split into genesis and persistence ($g$ and $p$). * indicates {\itshape Hydrogeminium natans} neighborhood kernel, other CA use {\itshape Orbium} kernel parameters \citep{chan2019}.
    \label{table:gliders} 
    }
\end{table}

We recovered gliders in 17 of 17 select CA available online\footnote{\url{https://chakazul.github.io/Lenia/JavaScript/Lenia.html}}, previously described in the Lenia framework . Table \ref{table:gliders} lists 5 examples each of Lenia CA and evolved CA which supported glider evolution.


Table \ref{table:metrics_1} includes metrics based on Eppstein's heuristics \citep{eppstein2010} and approximate CA classes \citep{wolfram1983, packard1985}. Mortality ratio is the proportion of grids with all zero cell values; fertility ratio is the proportion of grids where patterns escaped from a bounding box twice as tall and wide as the initialized area\footnote{Metrics based on 128 grids with 128 by 128 cells initialized with $\mathbb{U}(0,1)$ random uniform values.}. Putative classes are subjective, but reflect the diversity in the dynamics exhibited by these CA. The Lenia CA typically generate Turing pattern-like grid states, though not entirely static. s7 usually vanishes, s613, s643, and s11 exhibit chaotic, dynamics, and s113 generates a Turing pattern-type grid, similar to most Lenia CA. Despite different dynamics, all of these CA support mobile, self-organizing patterns. 

\begin{table}[h]
\center{
\begin{tabular}{c|c|c|c}\hline
Name & Fertility ratio & Mortality ratio & Putative class \\ \hline\hline

{\itshape Orbium} & 0.745/0.931 & 0.0377/0.0469 & II  \\
{\itshape D. valvatus} & 0.856/1.0 & 0.0/0.0 & II \\
{\itshape H. natans}* & 0.923/1.0 & 0.0/0.208 & II/IV \\
{\itshape P. s. labens} & 0.785/1.0 & 0.0/0.701 & II/IV \\
{\itshape S. valvatus} & 0.870/1.0 & 0.0/0.0 &  II \\

\hline

\textbf{s7}* & 0.0639/0.0138 & 0.577/0.976 & I \\
\textbf{s613}*  & 0.953/0.993 & 0.0/0.0 & III/IV \\
\textbf{s11}*  & 0.889/0.933 & 0.0212/0.0234 & IV \\
\textbf{s643}*  & 0.937/0.999 & 0.0/0.0  & III/IV \\
\textbf{s113}* & 0.693/0.992 & 0.00615/0.00781 & II \\
\end{tabular}
}
\caption{\label{table:metrics_1} Metrics and putative CA classes. Metrics are split (x/y) into first (x) and second (y) 512 CA steps, approximating initial and steady-state behavior.}
\end{table}

\begin{figure}
\begin{center}
  \includegraphics[width=0.4\textwidth]{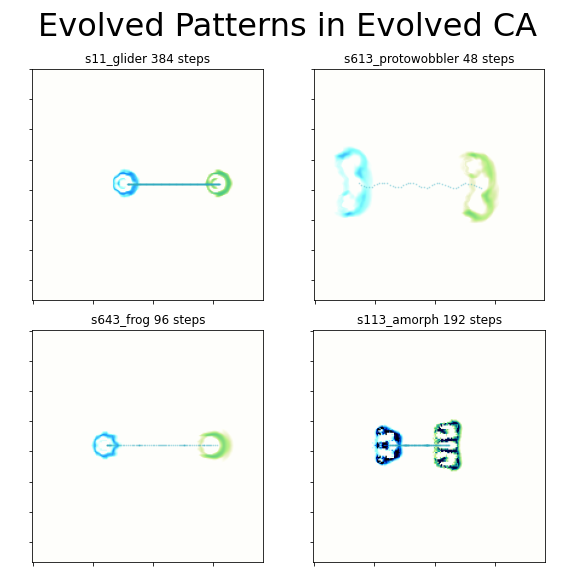}
  \caption{Evolved glider pattern trajectories in evolved CA. 
  \label{fig:evolved_patterns}
  }
\end{center}
\end{figure}

\section{Conclusions}

We demonstrated evolution of complex continuous CA, validating these by evolving gliders under the new rule sets. Selecting for poor halting prediction performance, and simply evolving CA rules that support both halting and persistent patterns, both yield CA that support gliders, the latter having lower computational and tuning overhead. 

Evolved CA supporting gliders in Table \ref{table:gliders} tend to have more diverse dynamics than their Lenia counterparts, and may help increase the already expansive diversity of bioreminiscent patterns discovered in Lenia, developed previously via manually and by interactive evolution.

\vspace{0.1cm}

\textbf{This work was supported by the National Science Foundation under the Emerging Frontiers in Research and Innovation (EFRI) program (EFMA-1830870).}
\footnotesize

\bibliographystyle{apalike}

\bibliography{references} 

\end{document}